\documentclass[11pt,reqno,a4,
]{amsart}

\usepackage{amsmath,amssymb}
\usepackage{graphicx}
\usepackage[pagebackref, colorlinks = true, linkcolor = blue, urlcolor  = blue, citecolor = red]{hyperref}

\usepackage[margin=1in]{geometry}

\usepackage{caption} 
\usepackage{subcaption}

\usepackage{url} 

\graphicspath{{figures_photos/}{figures_theory/}{figures_ex15/}}

\usepackage{layout}
\title{Central object segmentation by deep learning for fruits and other roundish objects}

\author{Motohisa Fukuda}
\address{MF: Yamagata University, Faculty of Science, 1-4-12 Kojirakawa, Yamagata, 990-8560 Japan}
\email{fukuda@sci.kj.yamagata-u.ac.jp}

\author{Takashi Okuno}
\address{TO: Yamagata University, Faculty of Science, 1-4-12 Kojirakawa, Yamagata, 990-8560 Japan}
\email{okuno@sci.kj.yamagata-u.ac.jp}

\author{Shinya Yuki}
\address{SY: Elix, Inc., Daini Togo Park Building 3F, 8-34 Yonbancho, Chiyoda-ku, Tokyo, 102-0081 Japan}
\email{shinya.yuki@elix-inc.com}

\keywords{
Deep Learning,
U-Net, 
Image Segmentation, 
Central Object,
Fruits 
}

\begin{document}
\maketitle

\begin{abstract}
We present \texttt{CROP} (Central Roundish Object Painter),
which  identifies and paints the object at the center of an RGB image.
Primarily \texttt{CROP} works for roundish fruits in various illumination conditions, but
surprisingly, it could also deal with images of other organic or inorganic materials,
or ones by optical and electron microscopes, although \texttt{CROP} was trained solely by 172 images of fruits.
The method involves image segmentation by deep learning, and 
the architecture of the neural network is a deeper version of the original \texttt{U-Net}.
This technique could provide us with a means of automatically collecting statistical data of fruit growth in farms.
As an example, we describe our experiment of processing 510 time series photos automatically to collect the data on the size and the position of the target fruit. 
Our trained neural network \texttt{CROP} and the above automatic programs are available on \texttt{GitHub}
(\url{https://github.com/MotohisaFukuda/CROP}),
with user-friendly interface programs.
\end{abstract}

\markright{\MakeUppercase{Central object segmentation by deep learning}}

\begin{figure}[htbp]
\begin{subfigure}{.24\textwidth}
  \centering
\includegraphics[width=1.0\linewidth]{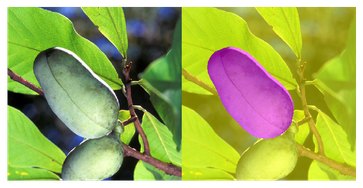}  
\caption{pawpaw.}
\label{fig:usda_pawpaw19a} 
\end{subfigure}
\begin{subfigure}{.24\textwidth}
  \centering
  \includegraphics[width=1.0\linewidth]{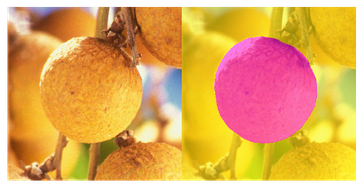}  
\caption{longan.} 
  \label{fig:usda_longan8a}
\end{subfigure}
\begin{subfigure}{.24\textwidth}
\centering
\includegraphics[width=1.0\linewidth]{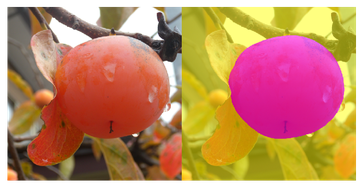}  
\caption{persimmon.}
\label{fig:murayama10}
\end{subfigure}
\begin{subfigure}{.24\textwidth}
  \centering
  \includegraphics[width=1.0\linewidth]{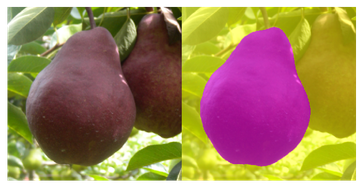}  
\caption{pear.}
\label{fig:murayama35a}
\end{subfigure}
\vspace{-2mm}
\caption{\texttt{CROP} identifies and paints central fruits of various kinds.
These images were not used for the training or validation, and can be considered as test data,
i.e. they are new to \texttt{CROP}. 
The original photos of Figure \ref{fig:usda_pawpaw19a} and \ref{fig:usda_longan8a}
are credited to USDA ARS (Scott Bauer), and  those of Figure \ref{fig:murayama10} and Figure \ref{fig:murayama35a} Hideki Murayama.} 
\label{fig:illumination}
\end{figure}

\newpage
%
\tableofcontents

\section{Introduction}

\subsection{Computer vision and fruit cultivation}\label{sec:cv}
Techniques of computer vision have various applications in fruit production,
for example making yield estimation by processing images of farms before the harvest.
Such pre-harvest estimate is crucial for efficient resource allocations 
and these computer vision techniques certainly
would constitute a part of smart agriculture \cite{review_detection}.

To this end, initially, algorithms of human-made feature extraction engineering
have been developed to locate fruits in camera images
\cite{grape_pixels, grape_detection, Sweetpepper_classical, mango_segmentation_classical, citrus_detection}, and
interestingly some used controlled illumination at night \cite{grape_night}.
Also, non-destructive size and volume measurement of fruits 
is useful in production and distribution of fruits \cite{review_nondestructive}. 
This is in fact one of our motivations in this paper. 
Naturally computer vision techniques have been developed, for example
using color analysis to conduct image segmentation \cite{mango_segmentation_color}.
Some used proper backgrounds for image segmentation
\cite{measurement_citrus, measurement_grape}, and even to 
develop a smart phone application \cite{measurement_SmartPhone}. 

Inevitably, \emph{machine learning} techniques came in to play,
for example for harvest estimation of tomatoes;
image segmentation based on pixels and blobs
and then \emph{X-means clustering} were applied \cite{tomatoML}. 
Even multi-spectral data was processed with \emph{conditional random field} for image segmentation for various crops  \cite{multispectral_segmentation_ML}.
In addition, for robotic harvesting, examples include the uses of \emph{Viola-Jones object detection framework} \cite{robotic_ml}, and
image segmentation with \emph{K-means clustering} \cite{robot_segmentation_ML}.
For accurate size measurement, \emph{support vector machine} performed image segmentation for apples with black backgrounds \cite{apple_measurement_ML} for example.

\subsection{Application of deep learning}\label{sec:deep}
\emph{Deep learning} is one of machine learning techniques.
There are more and more applications of deep learning not only in fruit production
but also in agriculture as a whole \cite{deepagri}.

Human-made algorithms require humans to set features to be extracted from raw data. 
In this case, the number of such features must be limited if one wants to establish algorithms within a reasonable time frame.
However, DNNs (Deep Neural Networks) extract important features automatically,
which is one of key properties of deep learning. 
Of course, DNNs process data in a black box,
but such data-driven methods could handle complex data, for example camera images in farms,
which are affected heavily by time of day, weather conditions, seasons, and
other illumination conditions such as shade and light reflection.
Therefore, huge amount of features would be necessary in processing such images,
and it would be simply too much for humans to find and list all important features in order to write algorithms by hand.

By contrast, things are completely different in processing data with DNNs, which consist of many layers.
Roughly writing, as data flows in a DNN, each layer makes the data little more abstract, so that
the DNN yields abstract understanding deep within itself. 
This way, humans do not have to pick important features by themselves. 
Interested readers can consult \cite{thebook} for more detailed theories of deep learning.

Recently one can find more and more applications of deep learning, for example counting apples \cite{apple_segmentation_DNN},
apples and oranges \cite{DL_countingAO}, and tomatoes \cite{DL_countingT}.
Also, in \cite{apple_segmentation_DNN_laplace} deep learning was applied to image segmentation in order to obtain horizontal diameters of apples,
by processing images taken on cloudy days or at dusk and pre-processed with Laplace transform.
Moreover in \cite{lightweight_cnn}, light-weighted neural networks were developed for image segmentation for agricultural robotics.

\texttt{Faster R-CNN} \cite{fasterRCNN} is one of famous DNNs for object detection,
and yielded successful applications for example in detecting apples \cite{FRCNN_apple},
 mangoes with spacial registration\cite{FRCNN_Space}, and sweet papers 
by using multi-modal data (color and Near-infrared)  \cite{FRCNN_multimodal}.
Despite its popularity, \texttt{Faster R-CNN} has disadvantages;
it gives rectangular bounding boxes for fruits in images but 
they do not give accurate size measurement.
By contrast, human-made algorithms for object detection have aspects of image (semantic) segmentation,
because they usually start with processing pixels to extract information such as colors and textures
before finally detecting fruits. 
This way, \emph{object detection} and \emph{image segmentation}  
are often  interwoven with each other in conventional techniques. 
In \cite{grapeMRCNN}, however, \emph{instance segmentation} via another DNN called \texttt{Mask-RCNN} \cite{maskrcnn} 
was used
 to overcome this disadvantage, together with space registration.
One can find detailed explanations on application of \texttt{Faster R-CNN} and  \texttt{Mask-RCNN} (and \texttt{YOLOv3} \cite{yolov3})
to images of grapes in \cite{grapeMRCNN}.

Our research put more weight on accurate measurement.
Our neural network, which we call \texttt{CROP} (Central Roundish Object Painter),  identifies and paints the fruit at the center of an image.
It works for roundish fruits in various illumination conditions.
This technique could provide us with a means of automatically collecting statistical data of fruit growth in farms,
and perhaps might allow automated robots to make better decisions. 
We also developed  programs to run \texttt{CROP} to automatically process time series photos by a fixed camera
to collect the data on the size and the position of the target fruit; see Section \ref{sec:timeseries}.
Our trained neural network \texttt{CROP} and the above automatic programs are available on \texttt{GitHub}
(\url{https://github.com/MotohisaFukuda/CROP}),
with user-friendly interface programs.

\section{Results}\label{sec:main}

\subsection{Datasets}\label{sec:datatrain}
In this research project, we used several groups of images from the internet and farms in Kaminoyama, Yamagata, Japan,
which are listed below with sample images.
\begin{description}
\item[Data\_Fruits] 172 images of a variety of fruits downloaded from (\url{https://pixabay.com}).\\
\includegraphics[width=1.0\linewidth]{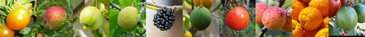} 
\vspace{0.1mm}

\item[Data\_Pears1] 26 images of pears in the farm in 2018 with \texttt{Brinno BCC100} (time-lapse mode).
\includegraphics[width=1.0\linewidth]{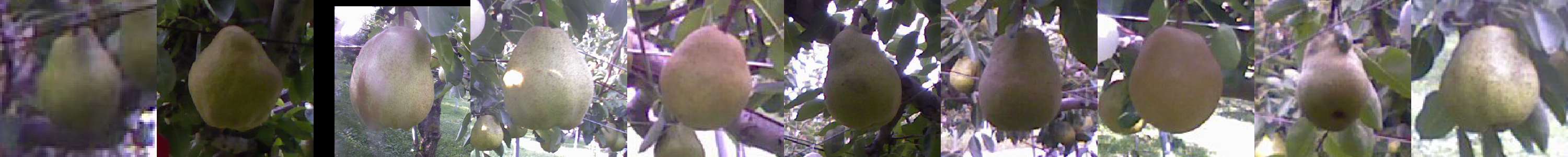} 
\vspace{0.1mm}

\item[Data\_Pears2] 86 images of pears in the farm in 2019 with various cameras.\\
\includegraphics[width=1.0\linewidth]{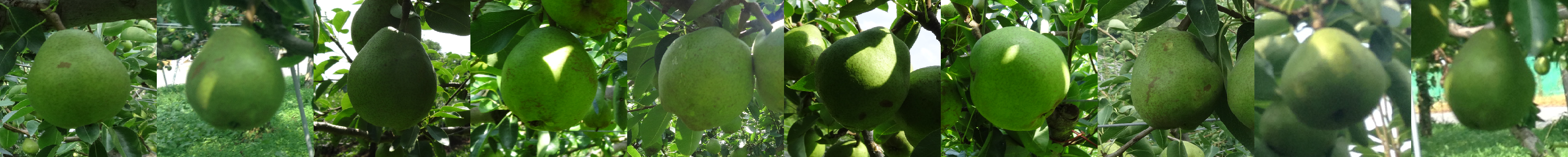} 
\end{description}
\texttt{Brinno BCC100} is a time-lapse camera used by one of the authors to keep track on growth of pears,
which give blurred images to be processed. 

These images were all annotated by \texttt{labelme} \cite{labelme2016} to train and evaluate \texttt{CROP}.  
In Section \ref{sec:quantitative} quantitative analysis was conducted;
we trained \texttt{CROP} with \texttt{Data\_Fruits} being training data (80\%) and validation data (20\%),
and afterwards fine-tuned it with \texttt{Data\_Pears2} being training data (80\%) and validation data (20\%).
On the other hand, \texttt{CROP} in Section \ref{sec:center} was trained by all the images of \texttt{Data\_Fruits}
to make qualitative analysis.

\subsection{Qualitative analysis}\label{sec:center}
In this section, we show how \texttt{CROP} works by using sample images. 
We used all of \texttt{Data\_Fruits} for training because we have a limited number of annotated images;
we decided when to stop the training by processing random images on the internet. 
The sample images in this section do not belong to \texttt{Data\_Fruits},
so that they can be thought of as test data, i.e. they are new to \texttt{CROP}.
Predictions by \texttt{CROP} are represented by mask images pasted onto the original (cropped) images;
red pixels for the central object and yellow for the rest.

In this section and the next, some of the original photos were provided by Hideki Murayama, or came from some institutions. In the latter case, they are credited in the captions, and the explanations of the acronyms of the institutions and the disclaimers are found in the section titled ``About photos in this paper''.

\begin{figure}[htbp]
\begin{subfigure}{.24\textwidth}
  \centering
  \includegraphics[width=1.0\linewidth]{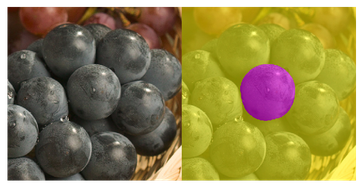}  
  \caption{the central grape.}
  \label{fig:usda_grape7a}
\end{subfigure}
\begin{subfigure}{.24\textwidth}
  \centering
  \includegraphics[width=1.0\linewidth]{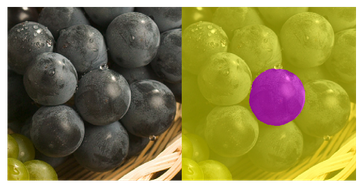}  
  \caption{another.}
  \label{fig:usda_grape7b}
\end{subfigure}
\begin{subfigure}{.24\textwidth}
  \centering
  \includegraphics[width=1.0\linewidth]{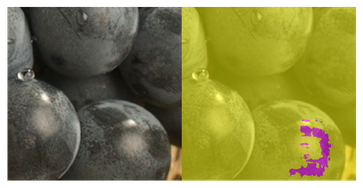}  
  \caption{confused.}
  \label{fig:usda_grape7c}
\end{subfigure}
\begin{subfigure}{.24\textwidth}
  \centering
  \includegraphics[width=1.0\linewidth]{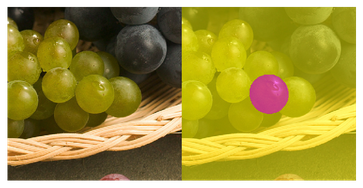}  
  \caption{wide-angle.}
  \label{fig:usda_grape7d}
\end{subfigure}
\caption{Identifying each central piece from bunches of grapes. The original photo is credited to USDA ARS (Peggy Greb). }
\label{fig:center1}
\end{figure}

First, the name of \texttt{CROP} does not come from the network architecture but from its function of
identifying and painting the object at the center of an image.
Examples in Figure \ref{fig:usda_grape7a} and Figure \ref{fig:usda_grape7b} indicate how 
\texttt{CROP} identifies individual grapes at the center of images.
By contrast, \texttt{CROP} got confused in Figure \ref{fig:usda_grape7c} because the central grape was behind others. 
On the other hand, it even could handle wide-angle images like in Figure \ref{fig:usda_grape7d}.

\begin{figure}[htbp]
\begin{subfigure}{.24\textwidth}
  \centering
  \includegraphics[width=1.0\linewidth]{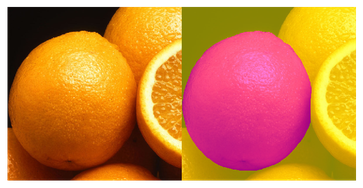} 
\caption{orange.}
  \label{fig:usda_orange3}
\end{subfigure}
\begin{subfigure}{.24\textwidth}
  \centering
  \includegraphics[width=1.0\linewidth]{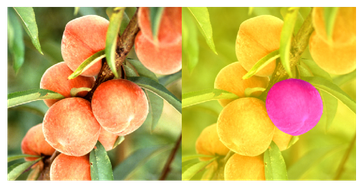}
\caption{peach.}
  \label{fig:usda_peach10b}
\end{subfigure}
\begin{subfigure}{.24\textwidth}
  \centering
  \includegraphics[width=1.0\linewidth]{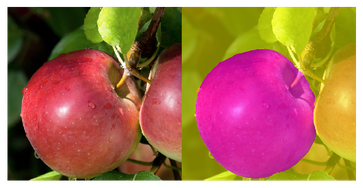} 
\caption{apple.} 
  \label{fig:usda_apple9}
\end{subfigure}
\begin{subfigure}{.24\textwidth}
  \centering
  \includegraphics[width=1.0\linewidth]{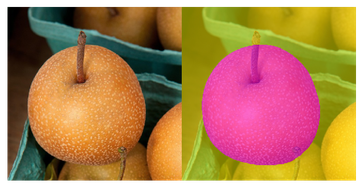}
  \caption{asian pear.}
  \label{fig:usda_asianpear11}
\end{subfigure}
\begin{subfigure}{.24\textwidth}
  \centering
  \includegraphics[width=1.0\linewidth]{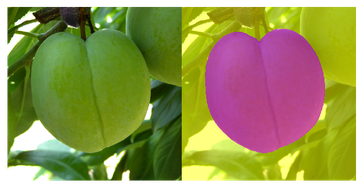}  
  \caption{apricot.}
  \label{fig:murayama24}
\end{subfigure}
\begin{subfigure}{.24\textwidth}
  \centering
  \includegraphics[width=1.0\linewidth]{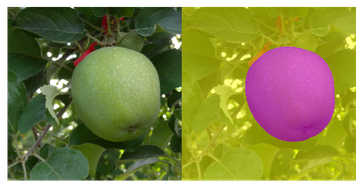} 
\caption{apple.}
  \label{fig:murayama14a}
\end{subfigure}
\begin{subfigure}{.24\textwidth}
  \centering
  \includegraphics[width=1.0\linewidth]{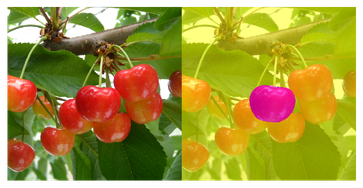} 
 \caption{cherry.}
  \label{fig:murayama21a}
\end{subfigure}
\begin{subfigure}{.24\textwidth}
  \centering
  \includegraphics[width=1.0\linewidth]{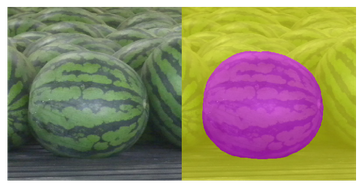}
\caption{watermelon.}
  \label{fig:murayama39a}
\end{subfigure}
\begin{subfigure}{.24\textwidth}
  \centering
  \includegraphics[width=1.0\linewidth]{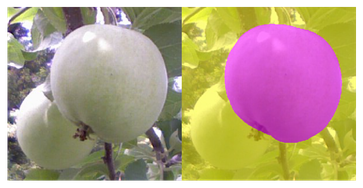} 
\caption{apple.}
  \label{fig:okuno_apple01_a}
\end{subfigure}
\begin{subfigure}{.24\textwidth}
  \centering
  \includegraphics[width=1.0\linewidth]{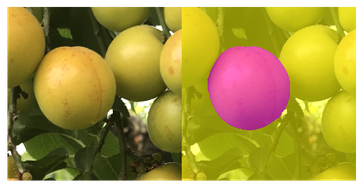} 
\caption{apricot.}
  \label{fig:okuno_apricot1589_a}
\end{subfigure}
\begin{subfigure}{.24\textwidth}
  \centering
  \includegraphics[width=1.0\linewidth]{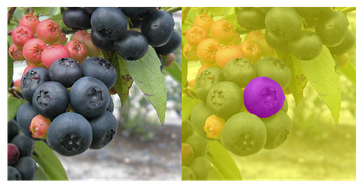} 
\caption{blueberry.}
  \label{fig:usda_blueberry15b}
\end{subfigure}
\begin{subfigure}{.24\textwidth}
  \centering
  \includegraphics[width=1.0\linewidth]{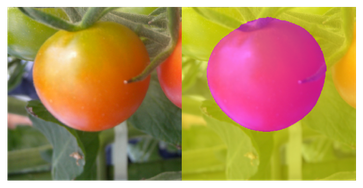} 
\caption{tomato.}
  \label{fig:murayama16_a}
\end{subfigure}
\caption{Examples. Some of the original photos are credited to USDA ARS; 
Figure \ref{fig:usda_apple9} and Figure \ref{fig:usda_asianpear11} to Peggy Greb,
Figure \ref{fig:usda_peach10b}  to Keith Weller,
Figure \ref{fig:usda_blueberry15b} to Mark Ehlenfeldt, and
Figure \ref{fig:usda_orange3} to unidentified persons. 
Moreover, 
Figure \ref{fig:murayama24}, Figure \ref{fig:murayama14a}, Figure \ref{fig:murayama21a},
Figiure \ref{fig:murayama39a}, Figure \ref{fig:murayama16_a} are credited to Hideki Murayama.}
\label{fig:successful}
\end{figure}

Let us go over some examples. Firstly, in Figure \ref{fig:successful}, one can see that 
\texttt{CROP} handled images of fruits of various shapes and colors.

\begin{figure}[htbp]
\begin{subfigure}{.24\textwidth}
  \centering
  \includegraphics[width=1.0\linewidth]{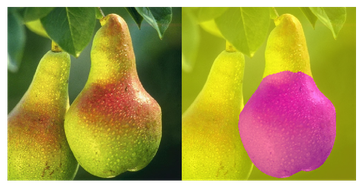} 
  \caption{Not round enough.}
  \label{fig:usda_pear13}
\end{subfigure}
\begin{subfigure}{.24\textwidth}
  \centering
  \includegraphics[width=1.0\linewidth]{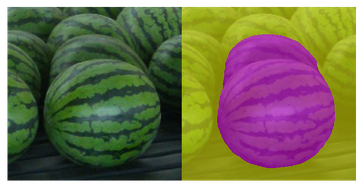}
  \caption{Mixed up.}
  \label{fig:murayama27b}
\end{subfigure}
\begin{subfigure}{.24\textwidth}
  \centering
  \includegraphics[width=1.0\linewidth]{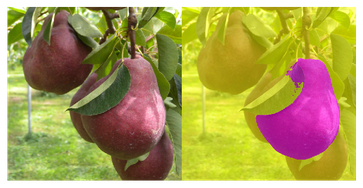}
  \caption{Disrupted by a leaf.}
  \label{fig:murayama35d}
\end{subfigure}
\begin{subfigure}{.24\textwidth}
  \centering
  \includegraphics[width=1.0\linewidth]{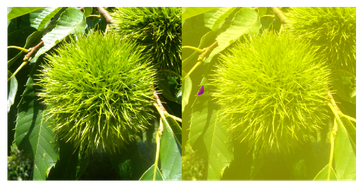} 
  \caption{Spiky boundaries.}
  \label{fig:murayama23}
\end{subfigure}
\caption{Unsuccessful examples. The original photo of Figure \ref{fig:usda_pear13} is credited to USDA ARS (Keith Weller).
Figures \ref{fig:murayama27b}, Figures \ref{fig:murayama35d} and Figures \ref{fig:murayama23} are credited to Hideki Murayama.}
\label{fig:unsuccessful}
\end{figure}

\begin{figure}[htbp]
\begin{subfigure}{.24\textwidth}
  \centering
  \includegraphics[width=1.0\linewidth]{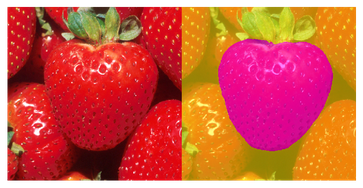}  
  \caption{}
  \label{fig:usda_strawberry12}
\end{subfigure}
\begin{subfigure}{.24\textwidth}
  \centering
   \includegraphics[width=1.0\linewidth]{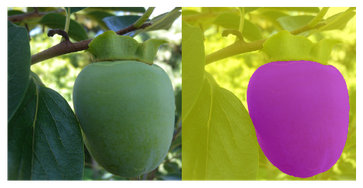} 
\caption{}
  \label{fig:murayama28}
\end{subfigure}
\begin{subfigure}{.24\textwidth}
  \centering
  \includegraphics[width=1.0\linewidth]{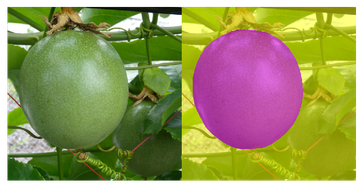} 
\caption{}
  \label{fig:murayama42}
\end{subfigure}
\begin{subfigure}{.24\textwidth}
  \centering
  \includegraphics[width=1.0\linewidth]{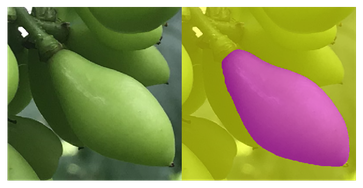} 
\caption{}
  \label{fig:okuno_grape1566_b}
\end{subfigure}
\caption{Peduncle and calyx. The original photo of Figure \ref{fig:usda_strawberry12} is credited to USDA ARS (Brian Prechtel).
Figure \ref{fig:murayama28} and Figure \ref{fig:murayama42} are credited to Hideki Murayama.}
\label{fig:peduncle_calyx}
\end{figure}

Secondly, we classified common mistakes by \texttt{CROP}, and included representative examples in Figure \ref{fig:unsuccessful}.
Here are our guesses why:
in Figure \ref{fig:usda_pear13}, the object is not round enough,
in Figure \ref{fig:murayama27b}  the angle of two meeting boundaries is not acute enough,
in Figure \ref{fig:murayama35d} there is a disruptive object, and in Figure \ref{fig:murayama23} `` there is no boundary''. 
Thirdly, however,
\texttt{CROP} can ignore those unimportant parts such as peduncle and calyx, like in Figure \ref{fig:peduncle_calyx}.

\subsection{Acquiring general ability}\label{sec:general}
\begin{figure}[htbp]
\begin{subfigure}{.24\textwidth}
  \centering
  \includegraphics[width=1.0\linewidth]{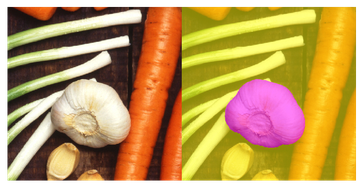} 
  \caption{a garlic.}
  \label{fig:usda_vegi20a}
\end{subfigure}
\begin{subfigure}{.24\textwidth}
  \centering
  \includegraphics[width=1.0\linewidth]{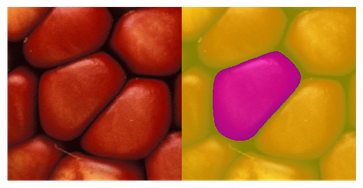} 
  \caption{a cell of corn.}
  \label{fig:usda_corn21a}
\end{subfigure}
\begin{subfigure}{.24\textwidth}
  \centering
  \includegraphics[width=1.0\linewidth]{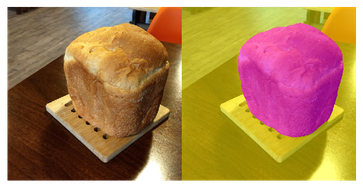} 
  \caption{a loaf of bread.}
  \label{fig:coffee1}
\end{subfigure}
\begin{subfigure}{.24\textwidth}
  \centering
  \includegraphics[width=1.0\linewidth]{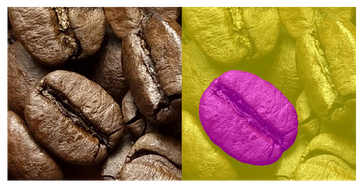} 
  \caption{a roasted coffee bean.}
  \label{fig:coffee10}
\end{subfigure}
\caption{Other foods. The original photos of Figure \ref{fig:usda_vegi20a} and Figure \ref{fig:usda_corn21a} are credited to USDA ARS (Scott Bauer and Keith Weller, respectively).}
\label{fig:foods}
\end{figure}

\begin{figure}[htbp]
\begin{subfigure}{.24\textwidth}
  \centering
  \includegraphics[width=1.0\linewidth]{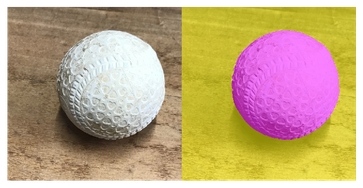} 
  \caption{a ball.}
  \label{fig:okuno_ball1556}
\end{subfigure}
  \begin{subfigure}{.24\textwidth}
  \centering
  \includegraphics[width=1.0\linewidth]{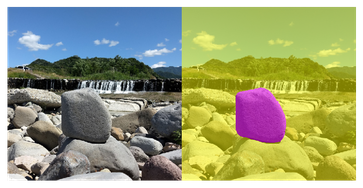} 
  \caption{a stone.}
  \label{fig:stone_br6}
\end{subfigure}
\begin{subfigure}{.24\textwidth}
  \centering
  \includegraphics[width=1.0\linewidth]{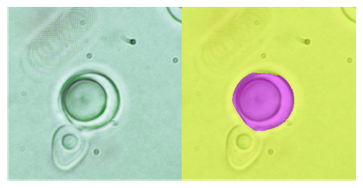}
\caption{a cell of yeast.}
  \label{fig:usda_yeast99b}
\end{subfigure}
\begin{subfigure}{.24\textwidth}
  \centering
  \includegraphics[width=1.0\linewidth]{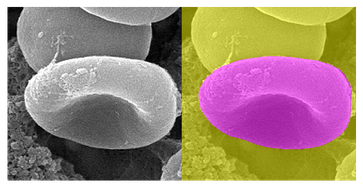}
\caption{a red blood cell.}
  \label{fig:cdc_7304_a}
\end{subfigure}
\caption{Various objects.
The original photo of Figure \ref{fig:usda_yeast99b} is credited to USDA ARS (Patricia Slininger), and 
that of Figure \ref{fig:cdc_7304_a} CDC (Janice Haney Carr).}
\label{fig:materials}
\end{figure}

\begin{figure}[htbp]
\begin{subfigure}{.24\textwidth}
  \centering
  \includegraphics[width=1.0\linewidth]{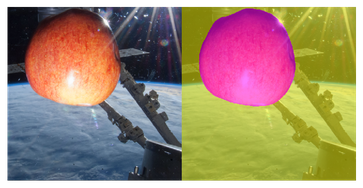}
  \caption{an apple in the ISS.}
  \label{fig:apple_nasa3}
\end{subfigure}
\begin{subfigure}{.24\textwidth}
  \centering
  \includegraphics[width=1.0\linewidth]{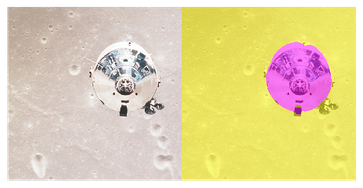} 
  \caption{Apollo 11.}
  \label{fig:apollo_nasa1}
\end{subfigure}
\begin{subfigure}{.24\textwidth}
  \centering
  \includegraphics[width=1.0\linewidth]{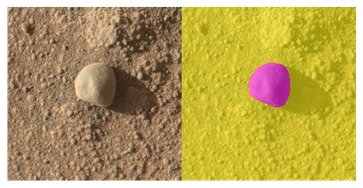}
  \caption{a rock on Mars.}
  \label{fig:rock_mars_nasa}
\end{subfigure}
\begin{subfigure}{.24\textwidth}
  \centering
  \includegraphics[width=1.0\linewidth]{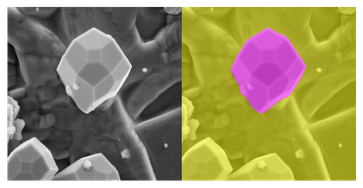} 
  \caption{a Moon iron crystal.}
  \label{fig:iron_moon_nasa1}
\end{subfigure}
\caption{The original photos of Figure \ref{fig:apple_nasa3} and \ref{fig:apollo_nasa1} are credited to NASA,
that of Figure \ref{fig:rock_mars_nasa} NASA/JPL-Caltech/MSSS, and that of Figure \ref{fig:iron_moon_nasa1} NASA/JSC.}
\label{fig:nasa}
\end{figure}

Although \texttt{CROP} has been trained solely by 172 fruit images (no transfer learning or fine-tuning),
it started to understand the meanings of boundaries of central roundish objects;
some other foods than fruits in Figure \ref{fig:foods}, various materials some of which were through microscopes in Figure \ref{fig:materials} , and photos in the space in Figure \ref{fig:nasa}. 

\section{Applying \texttt{CROP} in real pear farms}\label{sec:kaminoyama}
One of the authors actually plans to apply \texttt{CROP} in pear production. 
We argue in this section why it is difficult to predict harvest time of pears and 
how we could adapt \texttt{CROP} to images in real usage conditions,
so that one can collect time-series data of sizes of pears with fixed cameras in farms.
\subsection{Pear production in Japan}\label{sec:pear}
 ``La France''  is one of the most popular cultivars of European pear (Pyrus communis) in
Japan, its yield is about 70\% of European pears in Japan.
La France pomes are usually harvested at the mature-green stage and then chilled for stimulation of ethylene biosynthesis prior to being ripened at the room temperature. If the harvest is delayed or too early, the fruit does not ripe properly, and the texture, taste and flavor will be poor. In commercial pear cultivation, harvest time of La France greatly influences both the amount and quality of the harvest. Therefore, it is important to measure the maturity of the fruit precisely to optimize the time to harvest, but criteria to estimate the fruit maturity are limited such as fruit firmness and blooming date.
The fruit growth in terms of fruit size is described by an asymmetric sigmoid curve. The growth rate of the fruit on the tree is significantly affected by environmental factors and the physiologically active state. Precise time-lapse measurement of fruit growth should be useful for estimation of the fruit maturation status. In order to measure the fruit size change as it grows, the size of the same fruit must be measured repeatedly (daily) with a caliper. 

Research on sizes of agricultural products, without computer vision, was previously conducted 
in predicting optimum harvest time of carrots \cite{predict_optimumharvest}, estimating yield of pears \cite{predict_yield_pear} 
or anticipating cracking of bell peppers \cite{predict_cracking}.
Also devices for continuously measuring fruit size change have been developed, 
for example
the kiwifruit volume meter (KVM) \cite{instrument_kiwifruit},
stainless frames with potentiometers to be put on fruits \cite{instrument_low_cost}, and
flexible tapes around fruits to be read by infrared reflex sensors \cite{instrument_radial_tape}.
However, one of the authors plans to estimate fruit size based on images, possibly by using \texttt{CROP}.
We hope that
our research will improve the accuracy of fruit size measurement and reduce the labor required in the data collection.

\subsection{Getting fine-tuned for pears in farms}\label{sec:real}
Since \texttt{CROP} was initially trained by clear images of various fruits of \texttt{Data\_Fruits}, 
we used a technique called \emph{fine-tuning} to re-train \texttt{CROP} by  
rather unclear images of pears of \texttt{Data\_Pears2}, so that it can process similar images of \texttt{Data\_Pears1}.
One can consult Section \ref{sec:datatrain} for these datasets, and Section \ref{sec:quantitative} for quantitative analysis,
where the effect of fine-tuning is investigated. 
In the rest of the section, however, let us make qualitative analysis of fine-tuning through examples.

\begin{figure}[ht]
  \centering
 \begin{subfigure}{.49\textwidth}
  \centering
  \hspace{25mm} { before} \hspace{12mm} {after}
  \includegraphics[width=1.0\linewidth]{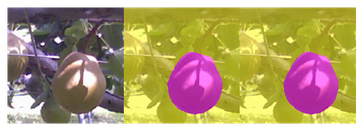} 
  \caption{good even before the fine-tuning.}
  \label{fig:22}
\end{subfigure} 
 \begin{subfigure}{.49\textwidth}
  \centering
  \hspace{25mm} {before} \hspace{12mm} { after}
  \includegraphics[width=1.0\linewidth]{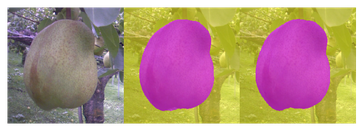}
  \caption{improvement at the bottom.}
  \label{fig:1}
\end{subfigure} 
\\
  \centering
 \begin{subfigure}{.49\textwidth}
  \centering
  \includegraphics[width=1.0\linewidth]{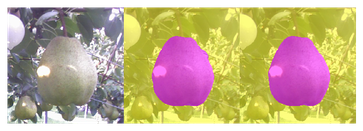} 
  \caption{improvement on the right border.}
  \label{fig:8}
\end{subfigure} 
 \begin{subfigure}{.49\textwidth}
  \centering
  \includegraphics[width=1.0\linewidth]{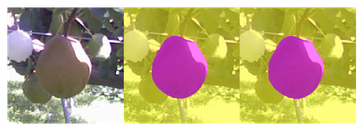}
  \caption{slight improvement at the top.}
  \label{fig:13}
\end{subfigure}
\\
  \centering
 \begin{subfigure}{.49\textwidth}
  \centering
  \includegraphics[width=1.0\linewidth]{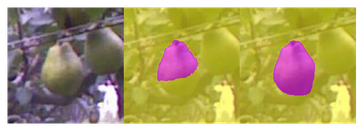} 
  \caption{dramatical improvement}
  \label{fig:18}
\end{subfigure} 
 \begin{subfigure}{.49\textwidth}
  \centering
  \includegraphics[width=1.0\linewidth]{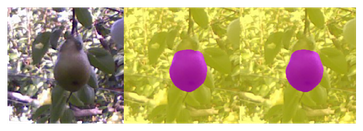}
  \caption{no improvement}
  \label{fig:5}
\end{subfigure}
  \caption{before and after fine-tuning.}
  \label{fig:realfarms}
\end{figure}

In Figure \ref{fig:realfarms}, 
each triple consists of the original image, and images processed before and after the fine-tuning, placed from left to right. 
Some images were processed well even before the fine-tuning, like in Figure \ref{fig:22}.
There are some small improvements in Figures \ref{fig:1}, \ref{fig:8}, \ref{fig:13}.
Some showed dramatical improvement, like in Figure \ref{fig:18}, 
and some on the contrary, like in Figure \ref{fig:5}.

\subsection{Application to time series photos}\label{sec:timeseries}
In this section, we show our way of using \texttt{CROP} in the local pear farms. All the implementations presented here are available on \texttt{GitHub}
(\url{https://github.com/MotohisaFukuda/CROP}). 
We will focus on our method of collecting data and have no intention to draw any scientific conclusions, although we make some remarks based on the data.
Indeed, eight photos were taken a day, at 8:00, 9:49, 11:49, 13:49, 15:49, 17:49, 19:49, 21:49, but the graphs and plots would look different if photos had been taken isochronously, for example once every three hours. In the following example, we processed 510 time series photos 
by a fixed camera to get the data on the size and the position of the chosen pear. All the process is automatic once we choose a target fruit in the first photo, and it took less than 14 minutes in this example. 
Note that these photos were taken in 2020 and are new to the \texttt{CROP} used in this section; see Section \ref{sec:datatrain}.

\subsubsection{Applying \texttt{CROP} to time series photos}
\begin{figure}[htbp]
\begin{subfigure}{.24\textwidth}
  \centering
  \vspace{10mm}
  \includegraphics[width=1.0\linewidth]{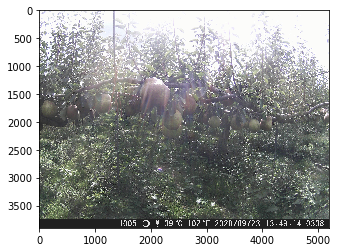} 
  \caption{}
  \label{fig:338_o}
\end{subfigure}
\begin{subfigure}{.24\textwidth}
  \centering
  \includegraphics[width=1.0\linewidth]{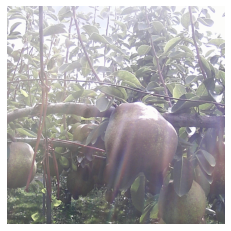} 
  \caption{}
  \label{fig:338_i}
\end{subfigure}
\begin{subfigure}{.24\textwidth}
  \centering
  \includegraphics[width=1.0\linewidth]{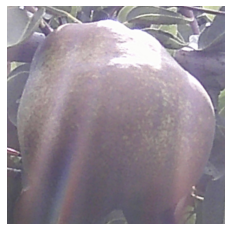}
  \caption{}
  \label{fig:338_a}
\end{subfigure}
\begin{subfigure}{.24\textwidth}
  \centering
  \includegraphics[width=1.0\linewidth]{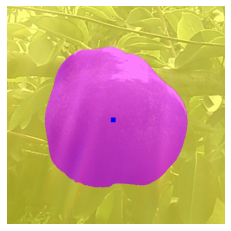}
  \caption{}
  \label{fig:338_f}
\end{subfigure}
\label{fig:process}
\caption{In the original photo of $5200\times3900$ pixels (Figure \ref{fig:338_o}), for example, choose an interested fruit (Figure \ref{fig:338_i}). Then, \texttt{CROP} will recognize the object (Figure \ref{fig:338_a}), and measure the pixel size and give the 2D-wise center of mass (Figure \ref{fig:338_f}). The photo id is 338.}
\end{figure}

Now, we explain about the newly developed programs which process time series photos automatically. 
The key ideas are: \texttt{CROP}
\begin{enumerate}
\item is able to detect central objects.
\item may be applied in different scales and one can pick the median of the measurement outcomes.
\item can keep track of the 2D-wise center of mass of the target.
\end{enumerate}
Let us elaborate the above three points one by one. 
First, once we specify an interested fruit by placing it around at the center of the photo frame (Figure \ref{fig:338_i}), then \texttt{CROP} can identify it, as in Figure \ref{fig:338_a}. By applying this functionality repeatedly, \texttt{CROP} can keep track of the fruit across time-series photos taken by a fixed camera.

\begin{figure}[htbp]
\begin{subfigure}{.6\textwidth}
  \centering
  \includegraphics[width=1.0\linewidth]{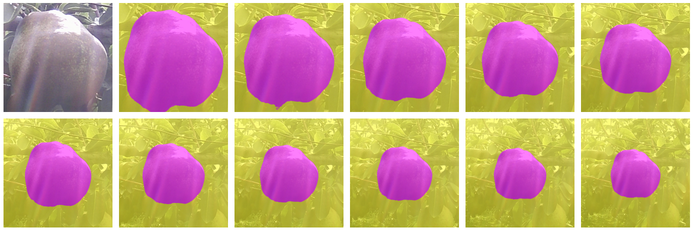}
  \caption{}
  \label{fig:338_tile}
\end{subfigure}
\begin{subfigure}{.38\textwidth}
  \centering
  \includegraphics[width=0.95\linewidth]{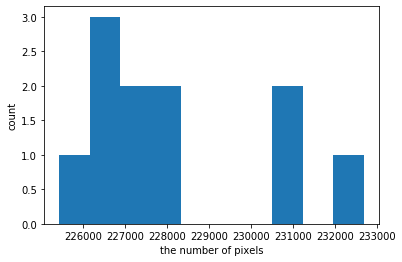}
  \caption{}
  \label{fig:338_h}
\end{subfigure}
\caption{\texttt{CROP} makes eleven measurements in different scale factors (Figure \ref{fig:338_tile}); from $\times 1.0$ to $\times 0.5$. The median of re-scaled values will be picked to exclude outliers; See the histogram in Figure \ref{fig:338_h}. The photo id is 338.}
\label{fig:median}
\end{figure}

Secondly, the fact that some of incorrect predictions by \texttt{CROP} largely depend on angles enables us to take the median of several measurement outcomes of different angles. Our implementation of this idea can be seen in Figure \ref{fig:median}, where inaccurate predictions in wider angles (Figure \ref{fig:338_tile}) appear as outliers in the histogram (Figure \ref{fig:338_h}). Note that all the numbers of pixels were re-scaled back to in the scale of the original photo before taking the median. This is why the programs output a decimal as the number of pixels. 

Thirdly, after choosing the best measurement in terms of median, one can also identify the center of mass in 2D photos (Figure \ref{fig:338_f}). We believe that this method is more reliable than using object detection algorithms. It is because errors in placing bounding boxes affect directly the positional data, but the center of mass is not so sensitive because pixel-wise mistakes will be averaged with other correct pixel-wise predictions. 

\subsubsection{Counting pixels and tracking the target}
Now we discuss how we processed at once 510 ($ 63 \times 8 + 5 + 1$) photos taken in Kaminoyama, Yamagata, Japan during 12 Aug 2020 13:49 --15 Oct 2020 8:00. Eight photos were taken a day, at 8:00, 9:49, 11:49, 13:49, 15:49, 17:49, 19:49, 21:49; the first three belong to the morning, the next two the afternoon, the last three the evening, in our terminology. No photos over night, unfortunately. The photos were given id's from 2 to 511 chronologically. These photos were taken by SC-MB68 a trail camera from SINEI.

\begin{figure}[htbp]
  \includegraphics[width=1.0\linewidth]{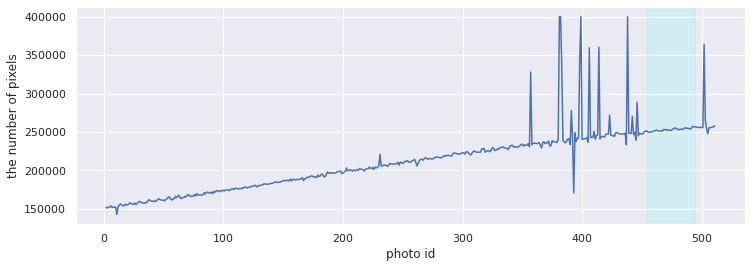} 
  \caption{The variation of the number of pixels during 12 Aug - 15 Oct 2020, predicted by \texttt{CROP}.  The variation during the five days highlighted by cyan color is found in Figure \ref{fig:pixels_seg}. Among possible outliers, four extremely large values are replaced by 400000. These images are found in Figure \ref{fig:outliers}.}
  \label{fig:pixels}
\end{figure}

\begin{figure}[htbp]
\begin{subfigure}{.24\textwidth}
  \centering
  \includegraphics[width=1.0\linewidth]{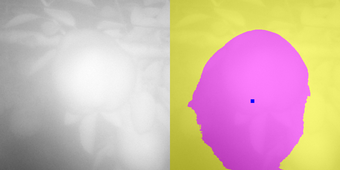} 
  \caption{381; 524382.2957}
  \label{fig:381_lite}
\end{subfigure}
\begin{subfigure}{.24\textwidth}
  \centering
  \includegraphics[width=1.0\linewidth]{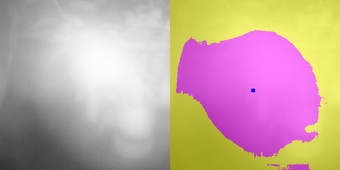}
  \caption{382; 935699.5221}
  \label{fig:382_lite}
\end{subfigure}
\begin{subfigure}{.24\textwidth}
  \centering
  \includegraphics[width=1.0\linewidth]{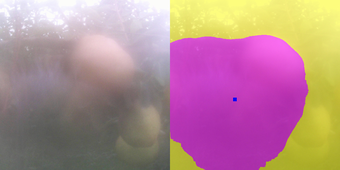} 
  \caption{399; 1135412.8060}
  \label{fig:399_lite}
\end{subfigure}
\begin{subfigure}{.24\textwidth}
  \centering
  \includegraphics[width=1.0\linewidth]{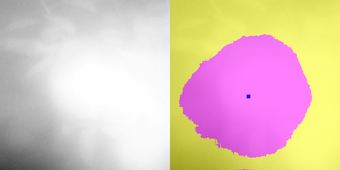}
  \caption{438; 444889.3609}
  \label{fig:438_lite}
\end{subfigure}
\caption{The four extreme outliers modified in Figure \ref{fig:pixels}. The measurements were completely messed up by mist. Each pair of numbers represent the photo id and the estimated number of pixels.}
\label{fig:outliers}
\end{figure}

First, Figure \ref{fig:pixels} shows the graph of the variation of the size of the target fruit estimated by \texttt{CROP}.  Unfortunately, there are terrible cases of miscounting; the worst four of which can be seen in Figure \ref{fig:outliers}, where it was heavily misty.

\begin{figure}[htbp]
  \includegraphics[width=1.0\linewidth]{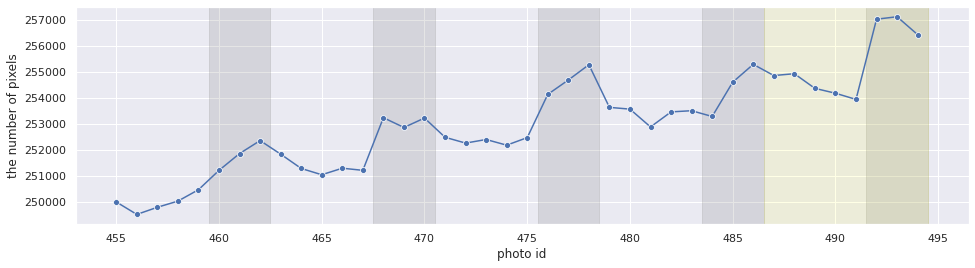} 
  \caption{The variation of the number of pixels during 08 - 12 Oct 2020, predicted by \texttt{CROP}. The darker background indicates the evening time. The eight photos of the day highlighted by yellow color are found in Figure \ref{fig:8tiles}. }
\label{fig:pixels_seg}
\end{figure}

\begin{figure}[htbp]
  \includegraphics[width=1.0\linewidth]{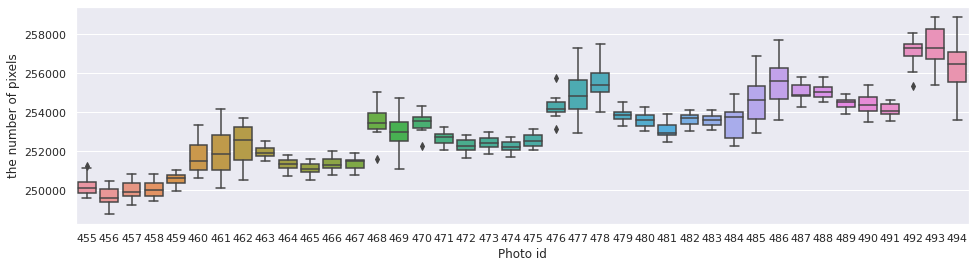} 
  \caption{Each distribution of the eleven outcomes of measurements (see Figure \ref{fig:338_tile}) is depicted by box plot during the period 08 - 12 Oct 2020. The variance is higher in the evening.
  The higher the variance is, the longer the box and the whisker are and the more individual points are.
  The evening time corresponds to 460--462, 468--470, 476--478, 484--486, 492--494.}
  \label{fig:measurements_seg}
\end{figure}

\begin{figure}[htbp]
  \includegraphics[width=1.0\linewidth]{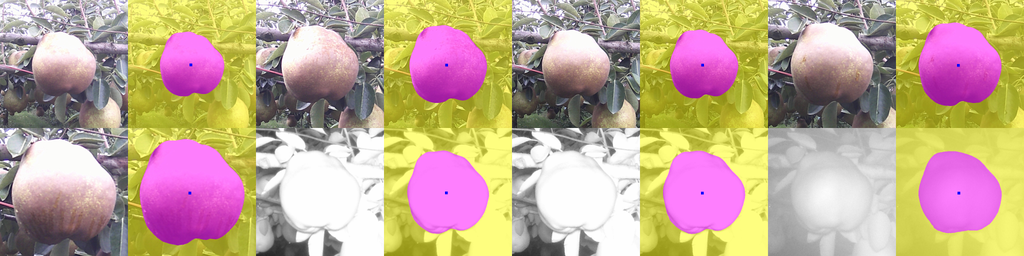} 
  \caption{Eight photos taken on 12 Oct 2020, whose photo id's are from 487 to 494.}
  \label{fig:8tiles}
\end{figure}

Next, to have a closer look, we focus on the five days (08--12  Oct  2020; photo id's are from 455 to 494) indicated by the highlight with cyan color in the graph.
In Figure \ref{fig:pixels_seg}, apparently
the fruit was larger in the evening, but we have to take account of optical effects by camera flash in the dark. Also,
the eleven measurement outcomes tend to have high variance in the evening, 
which can be identified in the box plot in Figure \ref{fig:measurements_seg}, by longer boxes and whiskers, and more individual points,  meaning that the data collected in the evening is less accurate.
Nevertheless, we may be able to claim that it grew over night based only on the data during the daytime.
Further, we focus on 12 Oct 2020, which corresponds to the yellow highlight in Figure \ref{fig:pixels_seg}. 
All the eight photos taken on the day and processed by \texttt{CROP} later are collected in Figure \ref{fig:8tiles}.

\begin{figure}[htbp]
\begin{subfigure}{.32\textwidth}
  \centering
  \includegraphics[width=1.0\linewidth]{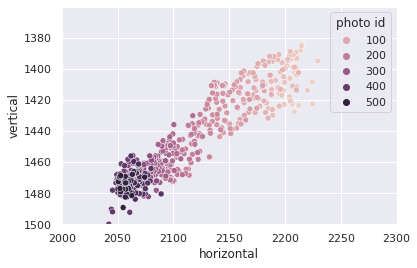} 
  \caption{The darker the later.}
  \label{fig:movement}
\end{subfigure}
\begin{subfigure}{.32\textwidth}
  \centering
  \includegraphics[width=1.0\linewidth]{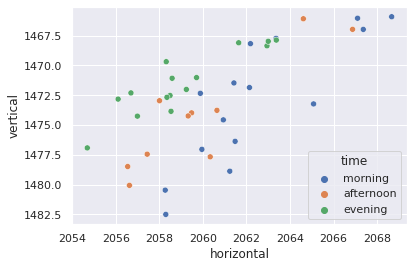}
  \caption{Different time slots.}
  \label{fig:movement_seg_dot}
\end{subfigure}
\begin{subfigure}{.32\textwidth}
  \centering
  \includegraphics[width=1.0\linewidth]{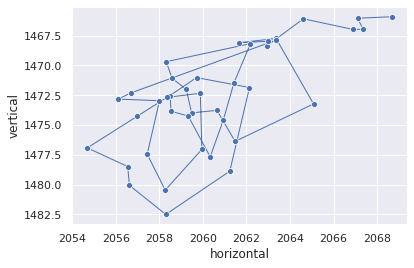}
  \caption{Trace of the movement.}
  \label{fig:movement_seg_line}
\end{subfigure}
\caption{The positions of the target fruit, predicted by \texttt{CROP}.
The coordinates are pixel numbers in the original photo, counting from the top-left corner; see Figure \ref{fig:338_o}. In Figure \ref{fig:movement} there are possible outliers inside and outside (to the below) the frame. Figure \ref{fig:movement_seg_dot} and Figure \ref{fig:movement_seg_line} correspond to the five days: 08 - 12 Oct 2020.}
\label{tba}
\end{figure}

Finally, one can see how the target fruit moved around during the whole season in Figure \ref{fig:movement}; outlies are also included.
Again, let's focus on the above five days. 
In Figure \ref{fig:movement_seg_dot}, the fruit seems to have hung rather higher in the evening. With Figure \ref{fig:movement_seg_dot}, one can trace the movement chronologically, based on the predictions by \texttt{CROP}.

\subsubsection{Remarks on technical matters}
It took only 791.1025 seconds (less than 14 minutes) to process 510 photos with NVIDIA TITAN Xp a GPU. All the process was automatic after specifying the fruit as in Figure \ref{fig:338_i}. During this process, for each photo, \texttt{CROP} made eleven measurements in different angles, stored the values as a csv file and all the mask images as thumbnails, chose the median and calculated the center of mass to save these data as a csv file and a PNG file. 
Without GPU it usually takes more than a minute to process just one photo, so use of a GPU is recommended for these new programs.

\section{Methodology and quantitative analysis}\label{sec:method}
In this section we make technical discussions.
Our choices of neural networks and loss functions are described in Section \ref{sec:network} and Section \ref{sec:loss}, respectively.
Then, in Section \ref{sec:quantitative},
quantitative analysis is made so that one can know how accurate predictions \texttt{CROP} could make for clear images,
and what are affects of fine-tuning (see Section \ref{sec:real} as well).
Also, Section \ref{sec:howto} explains how one could get rather stable predictions with \texttt{CROP} through the averaging process,
which was used in processing examples in this paper. 
Finally, we compare \texttt{CROP} with its original version of \text{U-Net} and
argue that the improvement probably depends on the deeper structure of \texttt{CROP}.

\subsection{Neural networks for image segmentation}\label{sec:network}
\emph{Image segmentation} can be seen as classification of image pixels.
For example, if you want to cut out a fruit in an image, 
all you have to do is classify each pixel into one of two classes: fruit and background. 
The number of classes can be more than one, and in this case
we want to classify each pixel into one of these classes. 
This kind of image process is called \emph{(image) semantic segmentation}.
In this sense, image semantic segmentation is much more difficult than 
\emph{object detection},
where one classifies each image, and not each pixel.
In the rest of this section we explain about our neural network and 
compare it with some other neural networks. 

\begin{figure}[thbp]
\begin{subfigure}{.49\textwidth}
  \centering
  \includegraphics[width=0.97\linewidth]{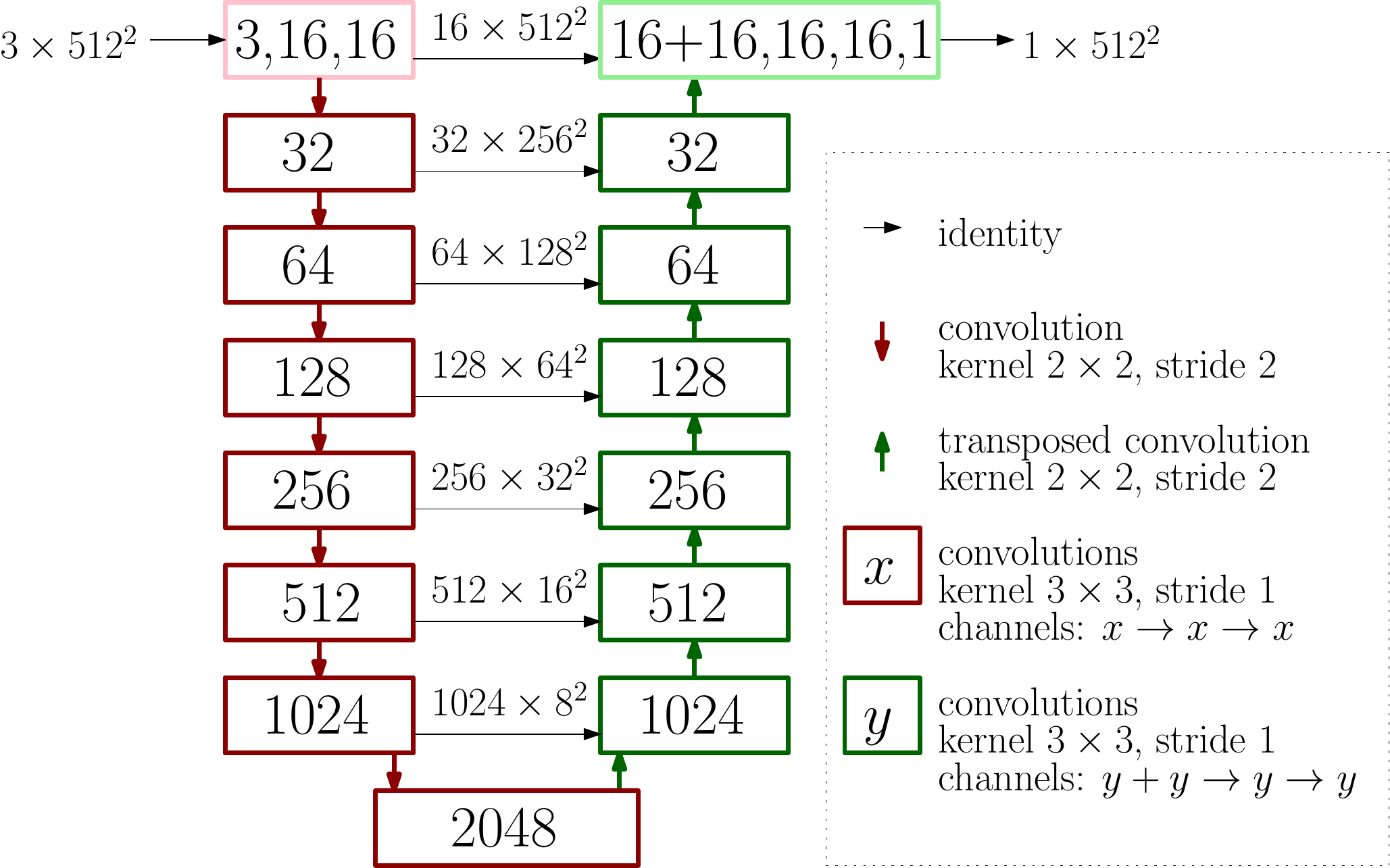}
  \caption{\texttt{CROP}.}
 \label{fig:ours}
\end{subfigure}
\begin{subfigure}{.49\textwidth}
  \centering
  \includegraphics[width=0.9\linewidth]{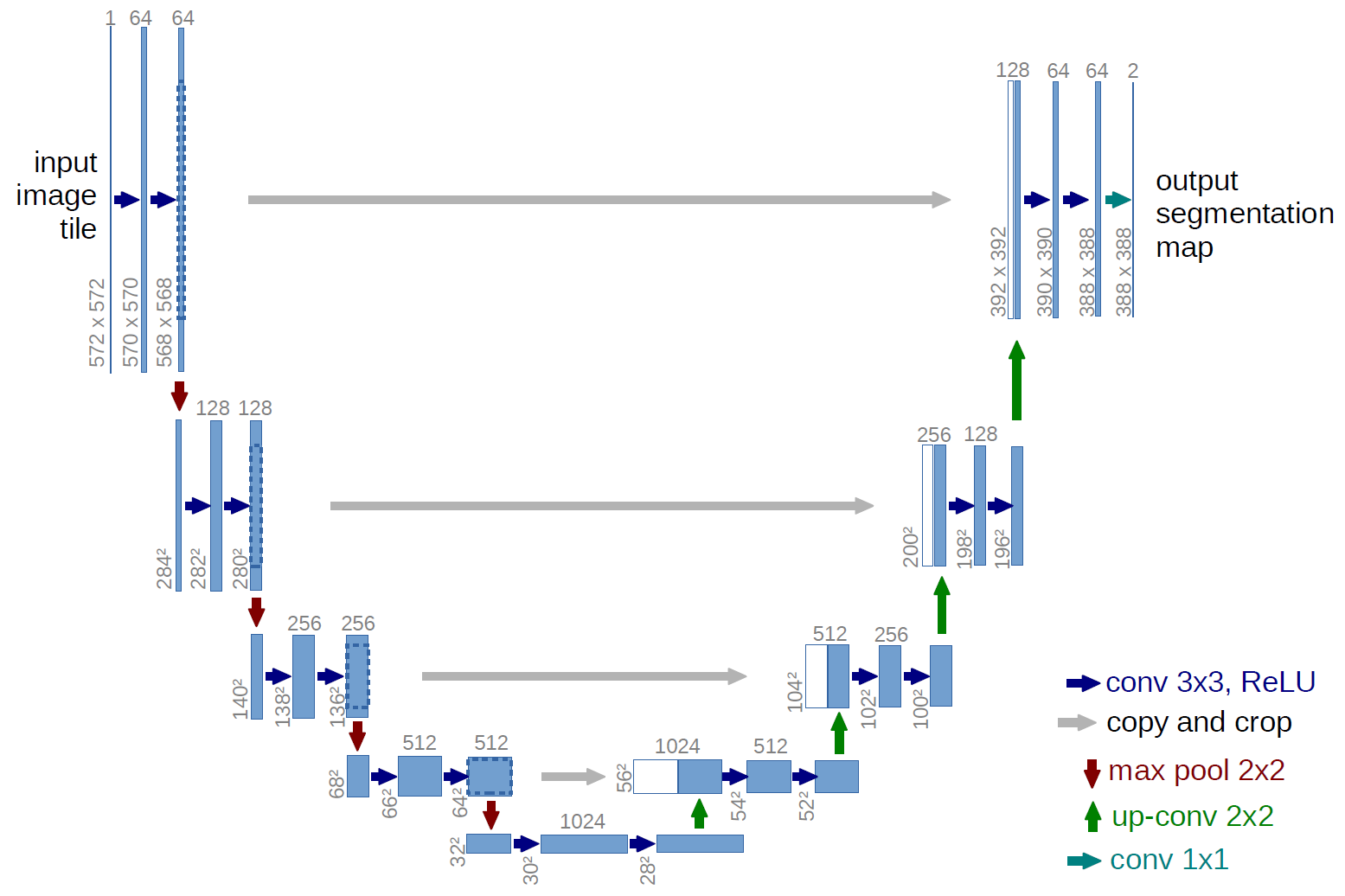}
  \caption{the original \texttt{U-Net}.}
  \label{fig:unet}
\end{subfigure}
\caption{comparison of \texttt{CROP} and the original \texttt{U-Net}.}
\label{fig:comparison_architecture}
\end{figure}

Strictly speaking, \texttt{CROP} is the name for our trained neural network for the specific purpose,
however, we call our neural network \texttt{CROP} even before the training to avoid confusion. 
As in Figure \ref{fig:ours}, inputs of \texttt{CROP}  
are RGB $512 \times 512$-pixel images and outputs $512 \times 512$-pixel mask images. 
The downward red arrows represent convolutions with kernel $2 \times 2$ and stride $2$,
which double the number of channels. 
Here, \emph{channel} corresponds to another dimension than height and width,
where the number of channels of RGB images is 3, and mask images 1. 
Similarly, the upward green arrows represent convolutions with kernel $2 \times 2$ and stride $2$,
which however make the number of channels half. 
The red rectangles are concatenations of two convolutions with kernel $3 \times 3$ and stride $1$,
which keep the number of channels unchanged.
The green rectangles are again concatenations of convolutions, where the second convolutions are the same as the ones from the red rectangles,
but the first ones are a bit different.
Their inputs are direct sums (in the channel space) of the outputs of the layers below and the copies from the left,
the latter of which are indicated by horizontal arrows. 
With these inputs then the first convolutions in the green rectangles make the number of channels half. 

To make our explanation complete, we explain the first and last boxes.
Pink and light green boxes represent again concatenations of convolutions with kernel $3 \times 3$ and stride $1$.
Through these layers the number of channels change as $3 \rightarrow 16 \rightarrow 16$ and $16 +16 \rightarrow 16 \rightarrow 16 \rightarrow  1$, respectively.
Note that \texttt{ReLU} and \emph{batch normalization} are applied adequately,
which are not explicit in the figure. 

The architecture of \texttt{CROP} is based on \texttt{U-Net} \cite{unet}.
\texttt{U-Net} was developed for medical image segmentation, and the architecture is depicted in Figure \ref{fig:unet},
which was taken from the website (\url{https://lmb.informatik.uni-freiburg.de}).
After the emergence of \texttt{U-Net},
many related research projects were conducted, including
\texttt{V-Net} for 3-dimensional medical images \cite{vnet},
from which we took the loss function for our training.

\texttt{U-Net} (and \texttt{V-net} as well) belongs to the family of \emph{Fully Convolutional Networks (FCN)},
which is a subset of \emph{Convolutional Neural Networks (CNN)}.
A CNN treats 2-dimensional data as it is, so that it gives better performance in image processing,
and mathematically it is literally comparable to convolution operation.
The idea of CNN is rather old  \cite{CNN} (some argue it comes from \cite{neocognitron}).
However, it proved useful in deep learning when \texttt{AlexNet} \cite{alexnet}
won  ImageNet Large Scale Visual Recognition Challenge in 2012, which is an image classification contest.
The key idea behind was that they combined the two ideas of CNN and Deep Neural Network (DNN).
This new architecture, realized by computational power of GPUs,
 enabled the neural network to improve the error rate dramatically. 

Naturally this architecture of (deep) FCN was then applied to the task of semantic segmentation in
 \cite{fullyconv_seg}, and with \emph{encoder-decoder} structure in \cite{deconv} and \cite{segnet}.
 The encoder-decoder structure consists of two parts;
 encoding with convolutions or max pools and decoding with transposed convolutions or up samples.
One can see this structure in the u-shaped part of \texttt{U-Net}.
In Figure \ref{fig:unet}, an input RGB image of $572 \times 572$ pixels becomes smaller 
by iterative application of max pools down to as small as $28 \times 28$,
although the number of channels becomes $1024$, i.e. $3 \times 527 \times 527$ pixels are transformed into $1024 \times 28 \times 28$ at the bottom,
where \texttt{U-Net} processes the data ``abstractly''. 
This down-sizing process corresponds to the $encoder$. 
By contrast, through the \emph{decoder} with up convolutions, \texttt{U-Net} yields an image of $388 \times 388$ with channel size one,
i.e. a mask image. 
Importantly, there are four \emph{skip connections}, which are represented by horizontal gray arrows in Figure \ref{fig:unet}.
They are supposed to transmit location information from the encoder to the decoder,
and this architecture characterizes \texttt{U-Net}.

Now,  \texttt{CROP} has a similar structure as in Figure \ref{fig:ours}.
The biggest difference is that the decoder and encoder are much deeper than those of \texttt{U-Net}.
\texttt{CROP} 
makes the size of an input image $2^7$ times smaller at the bottom, while \texttt{U-Net} nearly $2^4$ times smaller.
Those numbers $7$ and $4$ correspond to the number of red down arrows in Figure \ref{fig:ours} and Figure \ref{fig:unet}.
We believe that this difference in depth enables \texttt{CROP} to have more global and abstract understanding of images; 
see Section \ref{sec:depth} for quantitative experiments on this matter. 
Another difference is that
we adopted convolutions with kernel $2 \times 2$ and stride $2$ instead of max pools with kernel $2 \time 2$ in the decoder,
because convolutions learn but max pools do not. 
Note that  ``up convolution'' in Figure \ref{fig:unet} is same as ``transposed convolution'' in Figure \ref{fig:ours}.

Before concluding this section, we need to write about two more famous neural networks.
The first one is an implementation of \emph{instance segmentation} and is called \texttt{Mask R-CNN} \cite{maskrcnn}.
Instance segmentation can distinguish neighboring objects of the same class,
while semantic segmentation would mix them up because
it would just classify the pixels of these objects to one class. 
\texttt{Mask R-CNN}  was applied in \cite{grapeMRCNN}, to detect grapes bunch-wise. 
However, since we choose a fruit and fix a camera in the first place (Section \ref{sec:pear}),
we do not have to detect fruits and rather can focus on image segmentation.
The second one, for semantic segmentation, is called \texttt{DeepLabv3+} \cite{deeplabv3+}.
This neural network has \emph{atrous convolutions} to capture contextual information, 
and works very well for general purposes.
However, we do not need contextual information and would rather go for preciser segmentation ability of \texttt{U-Net},
which has been yielding many applications in medical image segmentation, where accuracy matters.

\subsection{Loss functions for image segmentation}
\label{sec:loss}Loss functions tell neural networks how to improve themselves.
In this sense, choice of loss function is very important in training neural networks. 
In this project, we picked \emph{soft dice loss} \cite{vnet}:
\begin{equation}\label{eq:sdl}
1 - 2 \left( \sum_{k} x_{k}t_{k} \right) \left/ \left( \sum_{k} x_{k}^2 + \sum_{k} t_{k}^2 \right) \right.
\end{equation}
where $k$ runs over all the pixels, $512 \times 512$ pixels in our case. 
Here, $\{x_k\}_k$ are outputs of \texttt{CROP} and $\{t_p\}_p$ are the target (``the right answers''). 
As usual $\{x_k\}_k$ take values between $0$ and $1$ after going through the sigmoid function: 
$1/(1+e^{-x})$, and $\{t_k\}_k$ exactly $0$ or $1$. 
As for the latter, $0$ and $1$ correspond to pixels of the background and the object, respectively,
and they are set based on the \emph{ground truth}, i.e. annotated data. 
Note that the loss vanishes if $x_k = t_k$ for all $k$'s.

Among loss functions, pixel-wise \emph{cross entropy}:
\begin{equation}\label{eq:ce}
\sum_{k} -  t_k \log x_k  - (1-t_k) \log (1-x_k)
\end{equation}
could seem to be a good choice if we consider image segmentation as classification of each pixel. 
In fact, the cross entropy loss is commonly used for image classification tasks. 
However, we did not use it because otherwise 
each pixel would carry the same share in \eqref{eq:ce}.
That is, in every batch, each mis-classified pixel 
makes the same amount of contribution to back-propagations (training process) regardless of the size of objects,
and as a consequence, the neural network could learn to rather ignore small objects. 
By contrast, with soft dice loss as in \eqref{eq:sdl}, 
such imbalance would be compensated by regularization,
which appears as the denominator. 
For similar reasons, we did not adopt $l_p$ loss ($p \geq 1$):
\begin{equation}
\sum_k (x_k - t_k)^p \ .
\end{equation}
Note that the above soft dice loss is a variant of \emph{dice loss},
This category of loss functions treat small and large objects relatively equally, and moreover if there are more than one classes
they also modify imbalance among different classes.  
In \cite{unet}, they applied more penalty to boundaries to segment images of cells, 
but we did not follow their path to avoid complication. 

Finally, \emph{IoU (Intersection over Union)}, or \emph{Jaccard index}:
\[
\text{IoU} (A,B) = \frac{|A \cap B|}{|A \cup B|} \ .
\]
measures how two sets are close to each other.
It takes the value between 0 and 1, corresponding to $A \cap B = \emptyset$ and $A=B$, respectively. 
This is not a loss function but an evaluation criteria, but shares the same spirit.

\subsection{Quantitative analysis of training process}\label{sec:quantitative}
In this subsection, we give data-scientific evaluation of the training process of \texttt{CROP},
by using soft dice loss and IoU.
All the evaluations were made with data augmentation unless otherwise stated.

\subsubsection{Training \texttt{CROP}}\label{sec:trainingCROP}
\begin{figure}[ht]
  \centering
 \begin{subfigure}{.45\textwidth}
  \centering
  \includegraphics[width=1.0\linewidth]{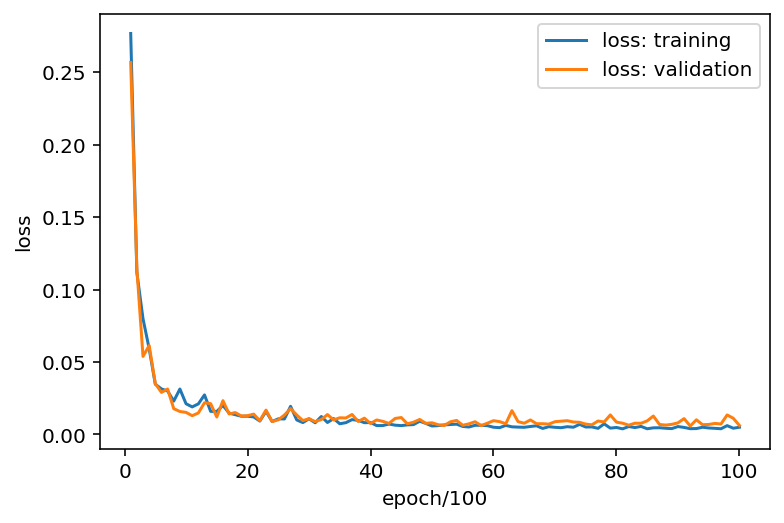} 
  \caption{loss}
\end{subfigure} 
 \begin{subfigure}{.45\textwidth}
  \centering
  \includegraphics[width=1.0\linewidth]{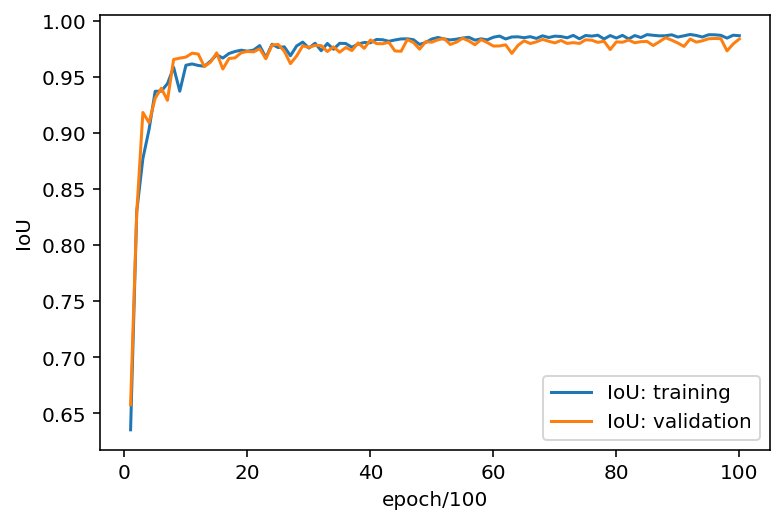}
  \caption{IoU}
\end{subfigure} 
\\
  \centering
 \begin{subfigure}{.45\textwidth}
  \centering
  \includegraphics[width=1.0\linewidth]{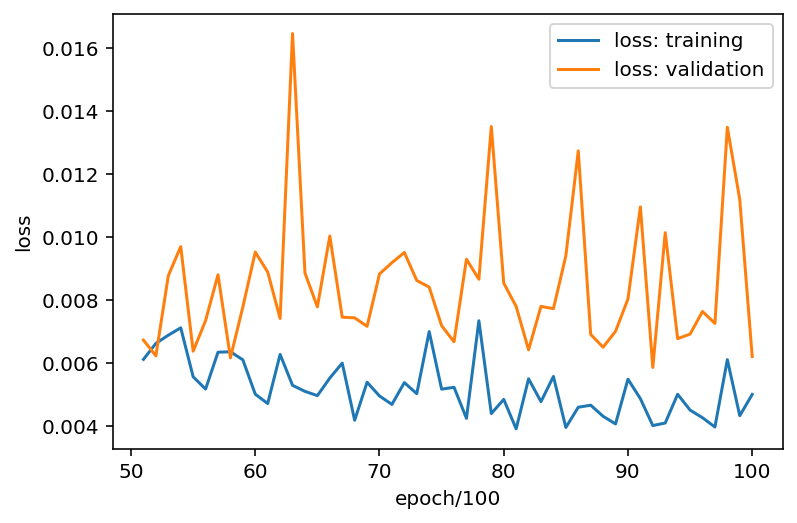} 
  \caption{loss for the last half}
\end{subfigure} 
 \begin{subfigure}{.45\textwidth}
  \centering
  \includegraphics[width=1.0\linewidth]{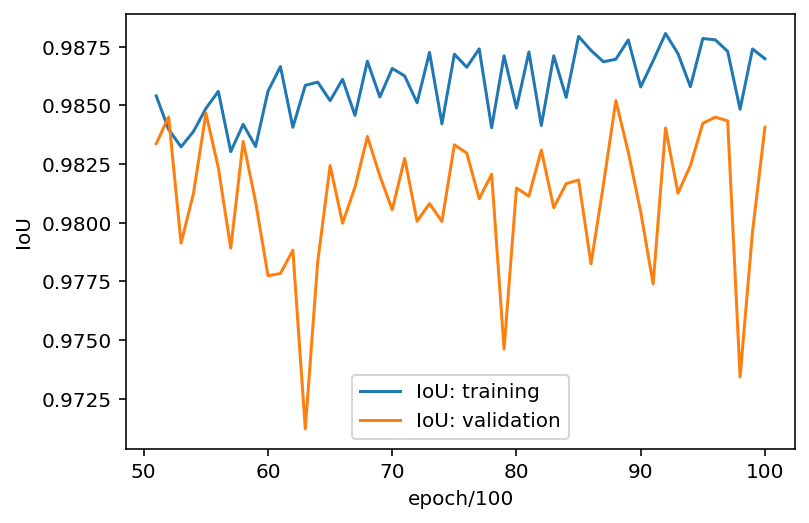}
  \caption{IoU for the last half}
\end{subfigure}
  \caption{training \texttt{CROP}
  with the training dataset (80\%: 137) and the validation dataset (20\%: 35) of \texttt{Data\_Fruits}, so that
  the best IoU for validation set was 0.985 at epoch 8,700.}
  \label{fig:training}
\end{figure}

For quantitative analysis we divided
\texttt{Data\_Fruits} into the training dataset (80\%: 137) and the validation dataset (20\%: 35).
Initial parameters of \texttt{CROP} were set randomly (we did not use pre-trained models), and 
were then ``optimized'' by \texttt{Adam} with learning rate $0.001$ and with batch size 14.
The result can be seen in Figure \ref{fig:training},
where the best IoU for the validation data was 0.985, achieved at the epoch 8,700.
To understand this value, suppose that we have a ground truth and a prediction of 100 pixels each, and that 99 pixels are correctly predicted.
Then, the IoU would be:
\[
\text{IoU(ground truth, prediction)} = \frac{99}{101} = 0.980...
\]
This optimal \texttt{CROP} was saved to the network dictionary named ``net\_dic\_0601\_08700' on \texttt{GitHub},
and was applied to examples in Section \ref{sec:real} to 
give predictions before the fine-tuning.

Note that \texttt{CROP} in Section \ref{sec:center} and Section \ref{sec:general} were
trained on all the images of \texttt{Data\_Fruits} and the training was stopped at epoch 5,000,
which seemed optimal through random samples on the internet.
The network dictionary is named ``net\_dic\_0314\_05000'' on \texttt{GitHub}. 

\subsubsection{Fine-tuning \texttt{CROP}}\label{sec:fine-tuning}
\begin{figure}[ht]
  \centering
 \begin{subfigure}{.45\textwidth}
  \centering
  \includegraphics[width=1.0\linewidth]{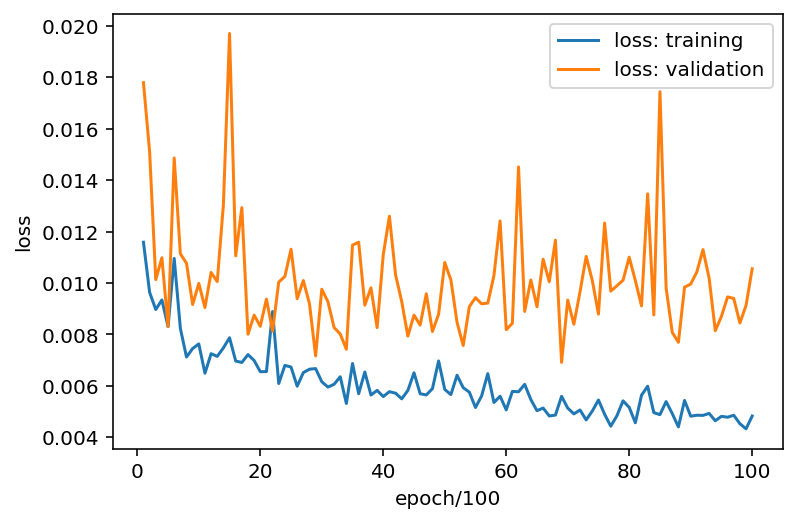} 
  \caption{loss}
\end{subfigure} 
 \begin{subfigure}{.45\textwidth}
  \centering
  \includegraphics[width=1.0\linewidth]{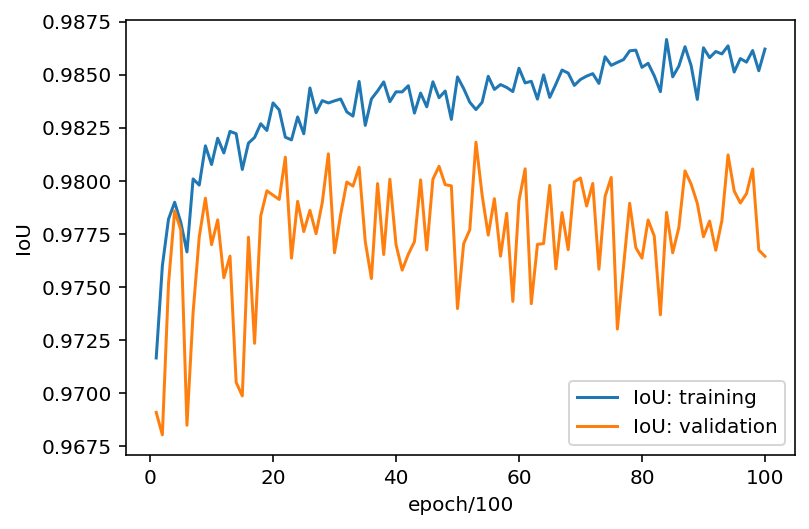}
  \caption{IoU}
\end{subfigure} 
  \caption{fine-tuning \texttt{CROP} with 
 the training dataset (80\%: 68) and the validation dataset (20\%: 18) of \texttt{Data\_Pears2}, so that
 the best IoU for validation set was  0.982, achieved at the epoch 5,200.}
  \label{fig:fine-tuning}
\end{figure}

For fine-tuning, we divided 
\texttt{Data\_Pears2} into the training dataset (80\%: 68) and the validation dataset (20\%: 18).
Then, we retrained the optimal \texttt{CROP} from Section \ref{sec:trainingCROP}, through
optimization by \texttt{Adam} with learning rate 0.0001 and with batch size 14.
One can see the process in Figure \ref{fig:fine-tuning}.
The best IoU for the validation data was 0.982, achieved at the epoch 5,200.
This optimal \texttt{CROP} was applied to examples in Section \ref{sec:real}.
This network dictionary is named ``net\_dic\_ft\_0601\_05200''  and placed on \texttt{GitHub}.
We also fine-tuned ``net\_dic\_0314\_05000'' for 5000 epochs to get 
the dictionary ``net\_dic\_ft\_0328\_1\_5000'', which was used in Section \ref{sec:timeseries} and is on \texttt{GitHub}.

Although, the improvement seems small from the graph in Figure \ref{fig:fine-tuning}, 
the IoU for \texttt{Data\_Pears1} improved a lot,
which can be seen on Table \ref{table:fine-tuning}, where
data augmentation was not applied 
because a good portion of the images in \texttt{Data\_Pears1} are already blur.

\begin{table}[htbp]
 \centering
 \begin{tabular}{|c|c|c|}
  \hline
   & before & after \\ \hline
IoU& 0.605 & 0.938 \\ \hline
\end{tabular}
\caption{IoUs for \texttt{Data\_Pears1} before and after the fine-tuning by \texttt{Data\_Pears2}.}
\label{table:fine-tuning}
\end{table}

\subsection{In using \texttt{CROP}}\label{sec:howto}
To make the predictions stable, we feed \texttt{CROP} eight different images made by applying 
actions of the dihedral group $D_4$ to the original input image.
They are eight different combinations of flips and rotations which keep a square unchanged in the two-dimensional space.
The final decision was made by averaging the corresponding eight predictions by \texttt{CROP}.
This is how we applied \texttt{CROP} to the examples in this paper,
 and it is implemented on \texttt{GitHub},
but the functionality can be switched off for quicker and possibly less accurate predictions.

\subsection{Discussions on depth of neural networks}\label{sec:depth}

\begin{figure}[ht]
  \centering
 \begin{subfigure}{.45\textwidth}
  \centering
  \includegraphics[width=1.0\linewidth]{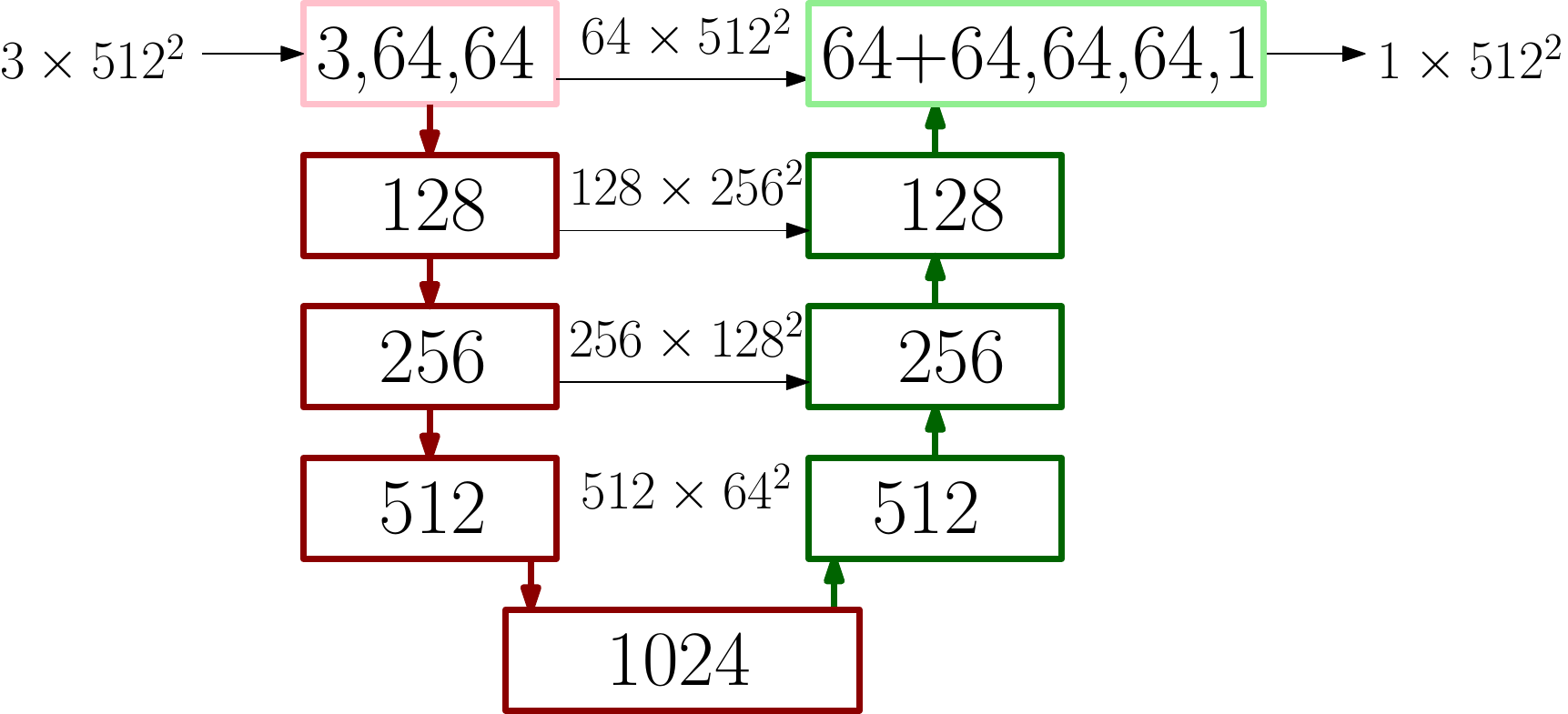} 
\end{subfigure} 
 \begin{subfigure}{.45\textwidth}
  \centering
  \includegraphics[width=0.4\linewidth]{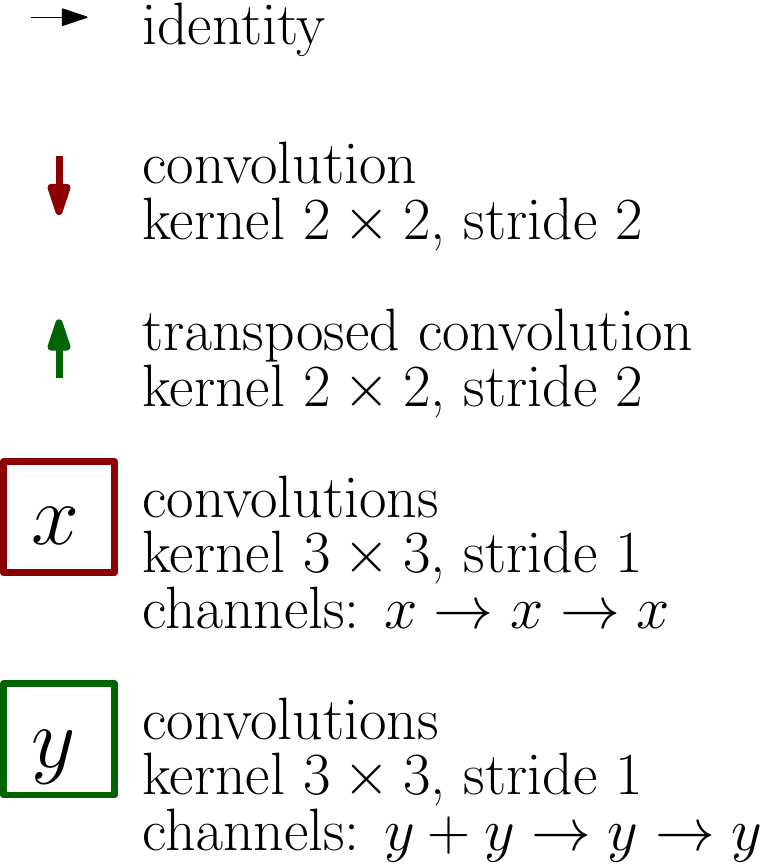}
\end{subfigure} 
  \caption{architecture of \texttt{CROP-Shallow}}
  \label{fig:net_shallow}
\end{figure}

In this section, we report on experiments on the shallow version of \texttt{CROP},
which is comparable to the original \texttt{U-Net}.
Provisionally we call it \texttt{CROP-Shallow} (see Figure \ref{fig:net_shallow}).
To this end, we trained \texttt{CROP-Shallow} and \texttt{CROP} in the same condition:
 loss function, optimization method, learning rate, but
random were partitions of training and validation sets, 
initialization of these neural networks, choice of batches and application of data augmentation. 
Also, the batch size was fixed to 6 for \texttt{CROP-Shallow}, which consumes a lot of GPU memory
probably because of the way how the channel size (or feature map size) changes.
However, the number of parameters itself is much smaller than that of \texttt{CROP}; 
40,103,873 for \texttt{CROP-Shallow} and 160,829,681 for \texttt{CROP}. 

\begin{figure}[ht]
\begin{subfigure}{.45\textwidth}
  \centering
  \includegraphics[width=1.0\linewidth]{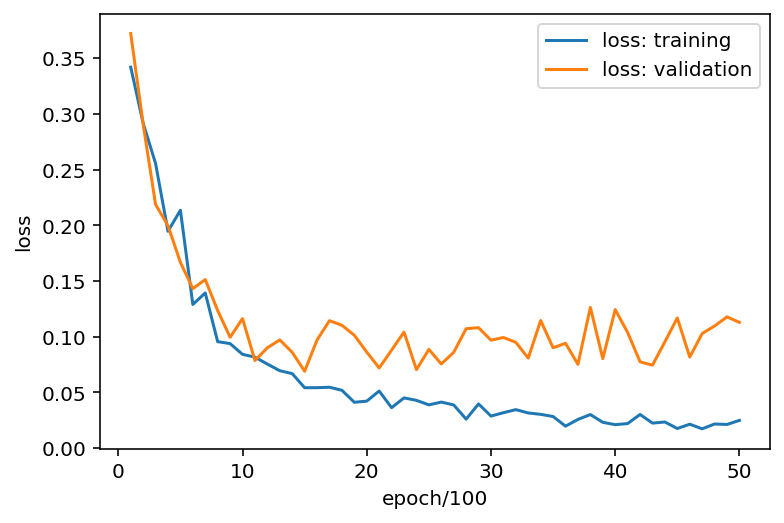} 
  \caption{loss}
  \label{fig:train_sopic_shallow}
\end{subfigure}
\begin{subfigure}{.45\textwidth}
  \centering
  \includegraphics[width=1.0\linewidth]{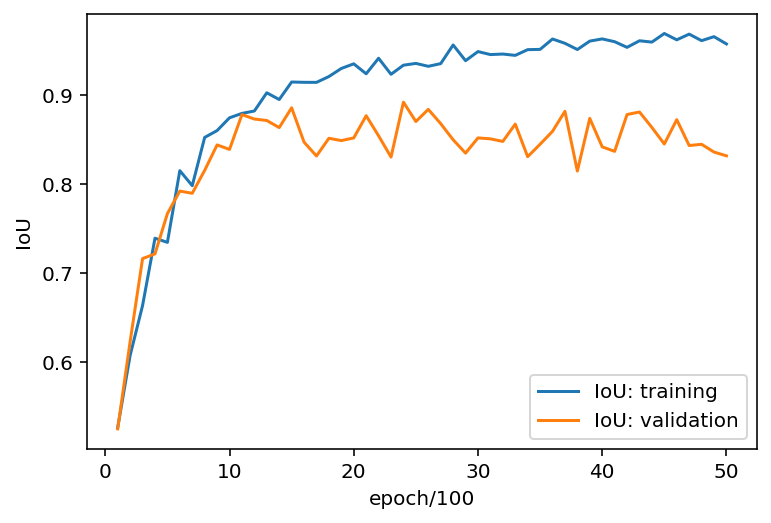}
  \caption{IoU} 
  \label{fig:train_deeper}
\end{subfigure}
\caption{
training \texttt{CROP-Shallow}
  with the training dataset (80\%: 137) and the validation dataset (20\%: 35) of \texttt{Data\_Fruits}, so that
  the best IoU for validation set was 0.899.}
\label{fig:trainingshallow}
\end{figure}

Despite these potential problems, however, we still believe that 
we captured significant difference between the two architectures.
See Figure \ref{fig:trainingshallow}, which shows over-fitting.
The difference can be read also from Table \ref{table:shallow}
of IoUs for the validation datasets.

\begin{table}[htbp]
 \centering
 \begin{tabular}{|c|c|c|}
  \hline
   & \texttt{CROP} &\texttt{CROP-Shallow} \\ \hline
best IoU& 0.985 & 0.899\\ \hline
\end{tabular}
\caption{the best IoUs for the validation datasets with \texttt{CROP} and \texttt{CROP-Shallow}
with the training dataset (80\%: 137) and the validation dataset (20\%: 35) of \texttt{Data\_Fruits},
split randomly for each.}
\label{table:shallow}
\end{table}

\section{Discussions on future directions}\label{sec:future}

As described above \texttt{CROP} was trained solely by 172 fruit images. 
With more training data, it would increase accuracy and generality. 
For accuracy, \texttt{CROP} comes with a disadvantage that it is heavy. 
It would need to slim down if it is to be used for harvesting robots, for example, where spontaneous data processing is required. 

In this project, we witnessed \texttt{CROP} had gained general ability to find the boundaries of  various types of objects at the center of images. 
However, it is different from \emph{salient object detection} \cite{salient}, 
as  \texttt{CROP} identifies a small central object like the one in Figure \ref{fig:usda_grape7d}. 
Perhaps, techniques developed in this paper could be applied 
to portrait segmentation \cite{portrait_auto, portrait_stylization, portrait_boundary}, 
or image segmentation for central objects in images, which we call \emph{central image segmentation}. 

\section*{About photos in this paper}
Example photos in this papers were provided by Motohisa Fukuda, Hideki Murayama and Takashi Okuno as well as the following institutions,
which are indicated by the acronyms. Original photos were cropped and processed by \texttt{CROP}.
\begin{itemize}
\item {\bf CDC:} Centers for Disease Control and Prevention (USA).
Use of Centers for Disease Control and Prevention (CDC) photos is not meant to serve as an official endorsement of any particular product or company, by the CDC, HHS, or the United States government.
\item {\bf NASA: } National Aeronautics and Space Administration (USA).
Use of photos of National Aeronautics and Space Administration (NASA) does not
state or imply the endorsement by NASA or by any NASA employee of a commercial product, service, or activity.
\item {\bf USDA ARS: } United States Department of Agriculture, Agricultural Research Service (USA).
Use of photos of Agricultural Research Service (ARS) of United States Department of Agriculture (USDA) is not meant to
infer or imply ARS endorsement of any product, company, or position.
\end{itemize}

\section*{Acknowledgement}
M.F. gratefully acknowledges the support of NVIDIA Corporation with the donation of the TITAN Xp GPU.
M.F. was financially supported by Leibniz Universit\"at Hannover
to present the result and have fruitful discussions in Hannover. 
M.F. thanks his colleague Richard Jordan for having a discussion with him to name our trained neural network and 
for suggesting better expressions in the title and abstract. 
M.F. and T.O. were financially supported by Yamagata University (YU-COE program).
T.O. thanks Kazumi Sato and  Yota Ozeki, who let him take images of pears in the farms,
and Yota Sato for annotating \texttt{Data\_Pears2}.
The authors also thank Kazunari Adachi of the engineering department for giving us valuable legal advice concerinig this research project,
and Hideki Murayama of the agricultural department for providing us with photos of fruits.

\bibliography{ref}{}
\bibliographystyle{alpha}

\end{document}